\documentclass[10pt, a4paper, conference, compsocconf]{IEEEtran}
\usepackage{amsmath}
\DeclareMathOperator{\Tr}{Tr}
\usepackage{cite}

\ifCLASSINFOpdf
   \usepackage[pdftex]{graphicx}
   \DeclareGraphicsExtensions{.pdf,.jpeg,.png}
\else
   \DeclareGraphicsExtensions{.eps}
\fi
\usepackage{nicefrac}

\begin{document}
%
\title{Color image quality assessment measure using multivariate generalized Gaussian distribution}


\author{\IEEEauthorblockN{Mounir Omari$^1$, Abdelkaher Ait Abdelouahad$^1$, Mohammed El Hassouni$^1$ and Hocine Cherifi$^2$}
\IEEEauthorblockA{$^1$LRIT URAC 29, University Mohammed V-Agdal\\
Avenue Ibn Battouta, B.P. 1014, Rabat, Morocco.\\
$^2$Laboratoire Electronique, Informatique et Image (Le2i) UMR 6306 CNRS\\
University of Burgundy, Dijon, France.\\
m.omari@fsr.um5a.ma, a.abdelkaher@gmail.com, mohamed.elhassouni@gmail.com, hocine.cherifi@u-bourgogne.fr}
}


%


\maketitle

\begin{abstract}
This paper deals with color image quality assessment in the reduced-reference framework based on natural scenes statistics.
In this context, we propose to model the statistics of the steerable pyramid coefficients by a Multivariate Generalized Gaussian distribution (MGGD).
This model allows taking into account the high correlation between
the components of the RGB color space. For each selected scale and orientation, we extract a parameter matrix from the three color components subbands.
In order to quantify the visual degradation, we use a closed-form of Kullback-Leibler Divergence (KLD) between two MGGDs.\\
Using \textquotedblleft{}TID 2008\textquotedblright{} benchmark, the proposed measure has been compared with the most influential methods according to the FRTV1 VQEG framework.
Results demonstrates its effectiveness for a great variety of distortion type. Among other benefits this measure uses only very little information about the original image.

\end{abstract}


%
\IEEEpeerreviewmaketitle

\section{Introduction}
\PARstart{O} {ver} the last years, we observe an exponential increase in the demand to evaluate the color image quality.
 To assess the performance of different image processing techniques, one has to measure the impact of the degradation induced by the processing in terms of perceived visual quality.
 For this purpose, subjective measures based essentially on human observer opinions have been introduced. These visual psychophysical judgments
 (detection, discrimination and preference) are made under controlled viewing conditions (fi{}xed lighting, viewing distance, etc.),
 generate highly reliable and repeatable data, and are used to optimize the design of imaging processing techniques. The test plan for
 subjective video quality assessment is well guided by Video Quality Experts Group (VQEG) including the test procedure and subjective
 data analysis. Objective image quality assessment models can be classified according to the information they use about the original image
 that is assumed to have perfect quality. Full-reference (FR) methods measure the deviation of the degraded image with the original one. In practice, they cannot be used when the reference image
 is not available. While FR methods are based on the knowledge of the reference image, no-reference (NR) methods are designed in order to grade
 the image quality independently of the reference. As they  are designed for one or a set of predefined specific distortion types
 they are useful only when the types of distortions between the reference and distorted images are fixed and known, and unlikely to be generalized.
 Reduced-Reference (RR) methods are designed to predict the perceptual quality of distorted images with only partial information about the reference images.
 Depending on the amount of information of the original image used both limiting cases (FR and NR) can be achieved.
 In practice the goal is to develop a RR metric using the minimum  amount of information about
 the reference because this side information must be always available in order to grade the quality and therefore it can be costly to transmit it.\\
 The structural similarity index (SSIM) \cite{key:2} and the visual signal to noise ratio (VSNR) \cite{key:4},
 are examples of successful full reference methods which have shown to be pretty effective in predicting the quality scores of human subjects.
 These conventional methods can be naturally extended to color images. Recently a reduced reference method based on the estimate of the structural similarity has been proposed by  Rehman\emph{ et al.}~\cite{key:6}.
 Another recent work of a reduced reference entropic difference method (RRED) has been proposed by Soundararajan \emph{et al.}~\cite{key:9}.
 The distortion measure was computed separately by the entropy difference of wavelet coefficients.
 This method has shown good performances in term of correlation with the human visual perception. However, a great quantity of information from the original image is needed in order to reach a high quality score.
 Indeed, the quantity of information extracted from the original image is a critical parameter of the reduced-reference schemes.
 Methods based on natural image statistics modeling  present a good trade-off between the quality score and the quantity of side information needed. The method proposed originally by Wang \emph{et al.} ~\cite{key:8} has been improved by Li \emph{et al.} ~\cite{key:10}.
 In this setting, a statistical model of steerable-pyramid coefficients (a redundant transform of wavelets family) is used.
 The distortion between the statistics of the original and the processed image is computed using the Kullback-Leibler divergence.
 These models can be naturally extended to color images using the same marginal distribution for each color component.
 However such an approach does not take into account the mutual information shared by the three color components.
In order to overcome this drawback, in this paper, we propose a solution that uses an estimate of the joint distribution of the three color components.
Unlike the RRED method, few information from the original image is needed to evaluate color image quality.
In the following, we concentrate on the RGB color space. This choice is motivated by its simplicity and its suitability to our model, however our approach can be extended to alternative color space models.
Note that the multivariate generalized Gaussian distribution has been proposed for RGB color texture modeling \cite{key:13}.
Based on this model, the parameters matrix are estimated by maximum likelihood method.
Finally, the proposed quality metric is computed by the Kullback-Leibler divergence between two MGGD models.\\
This paper is organized as {\normalsize follows.} Section 2 gives a brief review of the Multivariate generalized Gaussian distribution (MGGD).
 In section 3, we explain how we use this distribution to introduce the dependencies between the color components  and we present the distortion measure.
 Section 4  concerns the experimental results and finally a concluding remarks are presented in section 5.
\begin{figure*}
\centering
\includegraphics[width=6in]{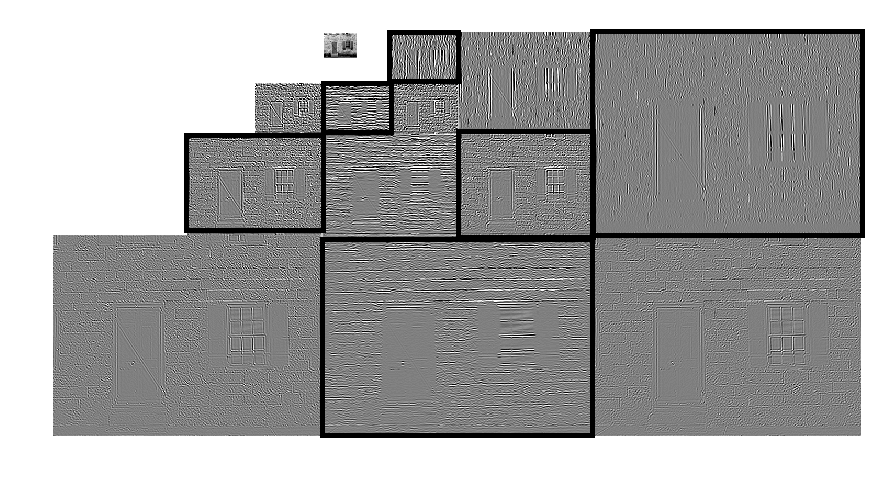}
\caption{Steerable pyramid decomposition of image. A set of selected subbands
(marked with boxes).}
\label{fig_sim}
\end{figure*}

\section{Multivariate Generalized Gaussian Distribution and Maximum likelihood estimation}
\subsection{Multivariate Generalized Gaussian Distribution}
In this work we consider a particular case of the Multivariate generalized gaussian distribution
introduced by Kotz~\cite{key:5}.
Upon multivariate extension of the univariate generalized Gaussian distribution a multivariate elliptical function does not exist.
Therefore, the MGGD is inspired from the univariate zero-mean GGD, which is described by the following expression:
\begin{equation} f(x|\alpha,\beta)=\frac{\beta}{2\alpha\Gamma[\nicefrac{1}{\beta}]}\exp[-(\nicefrac{|x|}{\alpha})^{\beta}],  \end{equation}
Where $\Gamma$ denotes the Gamma function, $\alpha$ is a
scale parameter and $\beta$ is the shape parameter.\\
The multivariate generalized Gaussian distribution is defined by the
following expression:
\begin{equation}
\begin{split}
 f(x|\Sigma,\beta)==\frac{\Gamma(\frac{m}{2})}{\pi^{\frac{m}{2}}\Gamma(\frac{m}{2\beta})2^{\frac{m}{2\beta}}}\frac{\beta}{\left|\Sigma\right|^{\frac{1}{2}}}\\
\star\exp\left\{ -\frac{1}{2}\left[x'\Sigma^{-1}x\right]^{\beta}\right\},
\end{split}
\end{equation}
Where $m$ is the dimensionality of the probability space ($m = 3$ for color space), $\beta$ is the shape parameter to control the peakedness of the distribution and the heaviness of its tails and $\Sigma$ is the covariance matrix. Note that,
the multivariate generalized Gaussian is also sometimes called the multivariate exponential power distribution.

\subsection{Maximum likelihood estimation}
We use the ML method for estimating the parameters of the MGGD. This involves setting the differential to zero of the logarithm of $f$ in (2).
Arranging the wavelet coefficients for a single subband in $n$ three-dimensional column vectors $x_{i}(i=1,.......,n)$, we get the following equations for $\Sigma$ and $\beta$, respectively.
\[ \Sigma=\frac{\beta}{n}\sum_{i=1}^{n}u^{\beta-1}x_{i}x{}_{i}^{'}, \]
\[ \sum_{i=1}^{n}\left\{\frac{\beta}{2}\ln\left(u\right)u^{\beta}-\frac{p}{2\beta}\left[\ln\left(2\right)+\Psi\left(\frac{p}{2\beta}\right)\right]-1\right\}=0, \]

Here, $\psi$ denotes the digamma function and $u$ is defined by $x'\Sigma^{-1}x$. These equations were solved recursively.
\section{Distortion measure}

At the receiver side, we use the features sent from the reference image to calculate the distortion measure.
In previous work, a number of authors have pointed out the relationship between KLD and MGGD
and used KLD to compare image statistics~\cite{key:13}.
In this paper, we use KLD to quantify the difference of MGGs distributions between a distorted image and a perfect quality reference image.
 We then examine how this quantity correlates with perceptual image quality for a wide range of distortion types.

\subsection{Kullback-Leibler divergence between two MGGD}

First we introduce the KLD between two distributions $p_{1}\left(x|\theta_{1}\right)$ and $p_{2}\left(x|\theta_{2}\right)$ that is denoted by
\begin{equation}
KLD\left(p_{1}||p_{2}\right)=\int p_{1}\left(x|\theta_{1}\right)\ln\frac{p_{1}\left(x|\theta_{1}\right)}{p_{2}\left(x|\theta_{2}\right)} dx,
\end{equation}
Do and Vettterli (2002)~\cite{key:11} have derived a closed-form expression
for the KLD between two univariate zero-mean generalized Gaussian.
 The KLD between two univariate generalized Gaussian distribution characterized by $\left(\beta_{1},\sigma_{1}\right)$ and $\left(\beta_{2},\sigma_{2}\right)$, where
the dispersions $\sigma_{i},i=1,2$ ($\Sigma_{i}$ reduced to $\sigma_{i}^2$ in the univariate case) is given by
\begin{equation}
\begin{split}
KLD\left(\beta_{1},\sigma_{1}||\beta_{2},\sigma_{2}\right)=\ln\left[\frac{\beta_{1}2^{\frac{1}{2\beta_{2}}}\sigma_{2}\Gamma\left(\frac{1}{2\beta_{2}}\right)}{\beta_{2}2^{\frac{1}{2\beta_{1}}}\sigma_{1}\Gamma\left(\frac{1}{2\beta_{1}}\right)}\right]+\\
\left(\frac{2^{\frac{1}{2\beta_{1}}}\sigma_{1}}{2^{\frac{1}{2\beta_{2}}}\sigma_{2}}\right)^{2\beta_{2}}\frac{\Gamma\left(\frac{2\beta_{2}+1}{2\beta_{1}}\right)}{\Gamma\left(\frac{1}{2\beta_{1}}\right)}-\frac{1}{2\beta_{1}}.
\end{split}
\end{equation}

When $\beta = 1$, the density reduces to the multivariate zero-mean Gaussian distribution.
In this case, the KLD between multivariate zero-mean Gaussian distributions is known since long (Kullback, 1968)~\cite{key:15}. It's given by
\begin{equation}
KLD\left(\Sigma_{1}||\Sigma_{2}\right)=\frac{1}{2}\left[\ln\frac{|\Sigma_{2}|}{|\Sigma_{1}|}+\Tr\left(\Sigma_{2}^{-1}\times\Sigma_{1}\right)-m\right].
\end{equation}

Where $\Sigma_{i}, i=1,2$ is the covariance matrix.

To our knowledge there is no analytic expression for the KLD between two multivariate zero-mean GGDs.
However, a closed form for the KLD between two bivariate zero-mean GGDs parameterized by $\left(\beta_{1},\Sigma_{1}\right)$ and $\left(\beta_{2},\Sigma_{2}\right)$
has been proposed (Verdoolaege and al, 2009)~\cite{key:16}. This quantity denoted $KLD\left(\beta_{2},\Sigma_{2}||\beta_{2},\Sigma_{2}\right)$ is defined by :
\begin{equation}
\begin{split}
KLD\left(\beta_{2},\Sigma_{2}||\beta_{2},\Sigma_{2}\right)=\ln \left[\frac{\Gamma\left(\frac{1}{\beta_{2}}\right)}{\Gamma\left(\frac{1}{\beta_{1}}\right)}2^{\left(\frac{1}{\beta_{2}}-\frac{1}{\beta_{1}}\right)}\left(\frac{|\Sigma_{2}|}{|\Sigma_{1}|}\right)^{\frac{1}{2}}\right]\\
+\ln \left[\frac{\beta_{1}}{\beta_{2}}\right]-\left(\frac{1}{\beta_{1}}\right)+2^{\frac{\beta_{2}}{\beta_{1}}-1}\frac{\Gamma\left(\frac{\beta_{2}+1}{\beta_{1}}\right)}{\Gamma\left(\frac{1}{\beta_{1}}\right)}\\
 \times\left(\frac{\gamma_{1}+\gamma_{2}}{2} \right)^{\beta_{2}} F_{1}\left(\frac{1-\beta_{2}}{2},\frac{-\beta_{2}}{2};1;A^{2} \right). 
 \end{split}
 \end{equation}
Where, $F_{1}(.,.;.;.)$ represents the Gauss hypergeometric function (Abramowitz and Stegun, 1965)~\cite{key:7}
which may be tabulated for $−1 < A < 1$ and for realistic
values of $\beta$. $\gamma_{i}=\left(\lambda^{i}_2\right)^{-1},i=1,2,$with $\lambda^{i}_2$ is the eigenvalues of $\Sigma_{2}^{-1}\Sigma_{1}$
while $A \equiv\frac{\gamma_{1}-\gamma_{2}}{\gamma_{1}+\gamma_{2}} $.

\begin{figure}
\centering
\includegraphics[width=3.5in]{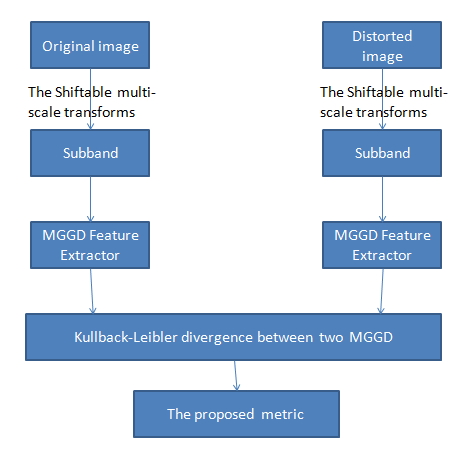}
\caption{Deployment of the reduced reference method with the multivariate generalized gaussian parameters.}
\label{fig_sim}
\end{figure}

\subsection{The proposed metric}

The Shiftable multi-scale transforms is represented with three orientations
and four scales to get twelve subbands. Here, we propose to exploit
conjointly the dependencies between the subbands and the dependencies between
the color components.
The Multivariate Generalized Gaussian Distribution (MGGD) is used in order
to model the joint statistics of color-image wavelet coefficients in  MGGD ($m=3$).
In the following, we will assume that $x$ in (2) is a matrix with the size $3 \times n$, where $n$ is the size of the subbands. So, automatically the size of $\Sigma$ is $3 \times 3$.

To take in consideration the dependencies between the subbands and minimize the number of parameters to be transmitted and the execution time, we choose, two orientations from each scale, as shown in Figure 1.\\
The diagram summarizing our approach is presented in figure 2. In the following we will consider $\beta=1$ to minimize the number of the parameters sent to the receiver side and we will use the KLD only between the covariance matrices $KLD\left(\Sigma_{1}||\Sigma_{2}\right)$.\\
Finally, when we get the difference between the six selected subbands, we use the following equation to combine the 6 values:
\begin{equation}
D=\text{\ensuremath{\sum}}_{i=1}^{6}D_{i}
\end{equation}
To produce an overall measure as follows:

\begin{equation}
Q=\log{_{2}}\left(1+\frac{1}{D_{0}}D\right)
\end{equation}

$D_{0}$ is a constant to control the magnitude of the distortion measure,
and it is equal to 0.1. The logarithmic function is involved
here to reduce the difference between a high values and low values of $D$, so that we can have values in
the same order.

\begin{table*}[!t]
\vspace{1.5ex}
 \centering %
\begin{tabular}{|c|c|c|c|c|c|c|}
\hline
{\scriptsize Distortion type }  & {\scriptsize $RRED_{16}^{M16}$~\cite{key:9} }  & {\scriptsize MSSIM~\cite{key:2} }  & {\scriptsize VIF~\cite{key:1} } & {\scriptsize PSNR }  & {\scriptsize WNISM~\cite{key:3} }  & {\scriptsize Our method}\tabularnewline
\hline
\hline
{\scriptsize Additive Gaussian noise}  & {\scriptsize 0.820}  & {\scriptsize 0.809}  & {\scriptsize 0.880}  & {\scriptsize 0.908}  & {\scriptsize 0.603}  & {\scriptsize 0.837}\tabularnewline
\hline
{\scriptsize Additive noise in}  & {\scriptsize 0.850}  & {\scriptsize 0.806}  & {\scriptsize 0.876}  & {\scriptsize 0.897}  & {\scriptsize 0.604}  & {\scriptsize 0.872}\tabularnewline
{\scriptsize color components is more intensive}  &  &  &  &  &  &  \tabularnewline
{\scriptsize than additive noise in the luminance component}  &  &  &  &  &  &  \tabularnewline
\hline
{\scriptsize Spatially correlated noise}  & {\scriptsize 0.842}  & {\scriptsize 0.820}  & {\scriptsize 0.870}  & {\scriptsize 0.917}  & {\scriptsize 0.599} & {\scriptsize 0.885}\tabularnewline
\hline
{\scriptsize Masked noise}  & {\scriptsize 0.833}  & {\scriptsize 0.816}  & {\scriptsize 0.868}  & {\scriptsize 0.851}  & {\scriptsize 0.633}  & {\scriptsize 0.942}\tabularnewline
\hline
{\scriptsize High frequency noise}  & {\scriptsize 0.901}  & {\scriptsize 0.869}  & {\scriptsize 0.908}  & {\scriptsize 0.927}  & {\scriptsize 0.908}  & {\scriptsize 0.909}\tabularnewline
\hline
{\scriptsize Impulse noise}  & {\scriptsize 0.741}  & {\scriptsize 0.687}  & {\scriptsize 0.833}  & {\scriptsize 0.872}  & {\scriptsize 0.593}  & {\scriptsize 0.839}\tabularnewline
\hline
{\scriptsize Quantization noise}  & {\scriptsize 0.831}  & {\scriptsize 0.854}  & {\scriptsize 0.780}  & {\scriptsize 0.870}  & {\scriptsize 0.619}  & {\scriptsize 0.859}\tabularnewline
\hline
{\scriptsize Gaussian blur}  & {\scriptsize 0.957}  & {\scriptsize 0.961}  & {\scriptsize 0.954}  & {\scriptsize 0.870}  & {\scriptsize 0.871}  & {\scriptsize 0.930}\tabularnewline
\hline
{\scriptsize Image denoising}  & {\scriptsize 0.949}  & {\scriptsize 0.957}  & {\scriptsize 0.916}  & {\scriptsize 0.941}  & {\scriptsize 0.864}  & {\scriptsize 0.942}\tabularnewline
\hline
{\scriptsize JPEG compression}  & {\scriptsize 0.933}  & {\scriptsize 0.935}  & {\scriptsize 0.917}  & {\scriptsize 0.873}  & {\scriptsize 0.834}  & {\scriptsize 0.842}\tabularnewline
\hline
{\scriptsize JPEG2000 compression}  & {\scriptsize 0.968}  & {\scriptsize 0.974}  & {\scriptsize 0.971}  & {\scriptsize 0.813}  & {\scriptsize 0.935}  & {\scriptsize 0.932}\tabularnewline
\hline
{\scriptsize JPEG transmission errors}  & {\scriptsize 0.870}  & {\scriptsize 0.874}  & {\scriptsize 0.859}  & {\scriptsize 0.751}  & {\scriptsize 0.875}  & {\scriptsize 0.862}\tabularnewline
\hline
{\scriptsize JPEG2000 transmission errors}  & {\scriptsize 0.742}  & {\scriptsize 0.853}  & {\scriptsize 0.850}  & {\scriptsize 0.831}  & {\scriptsize 0.691}  & {\scriptsize 0.867}\tabularnewline
\hline
{\scriptsize Non eccentricity pattern noise}  & {\scriptsize 0.713}  & {\scriptsize 0.734}  & {\scriptsize 0.762}  & {\scriptsize 0.581}  & {\scriptsize 0.452}  & {\scriptsize 0.746}\tabularnewline
\hline
{\scriptsize Local block-wise distortions of different intensity}  & {\scriptsize 0.824}  & {\scriptsize 0.762}  & {\scriptsize 0.832}  & {\scriptsize 0.617}  & {\scriptsize 0.590} & {\scriptsize 0.819}\tabularnewline
\hline
{\scriptsize Mean shift (intensity shift)}  & {\scriptsize 0.538}  & {\scriptsize 0.737}  & {\scriptsize 0.510}  & {\scriptsize 0.694}  & {\scriptsize 0.292}  & {\scriptsize 0.619}\tabularnewline
\hline
{\scriptsize Contrast change}  & {\scriptsize 0.542}  & {\scriptsize 0.640}  & {\scriptsize 0.819}  & {\scriptsize 0.587}  & {\scriptsize 0.701}  & {\scriptsize 0.662}\tabularnewline
\hline
{\scriptsize No.of scalars} & {\scriptsize L/36} & {\scriptsize L} & {\scriptsize L} & {\scriptsize L} & {\scriptsize 18} & {\scriptsize 54}\tabularnewline
\hline
\end{tabular}\caption{The Spearman rank correlation coefficient (SRCC) for TID 2008 database.\label{ls:tab}}

\vspace{1.5ex}

\end{table*}

\begin{figure}[htb]
\vspace{1.5ex}
 \centering \includegraphics[scale=0.6]{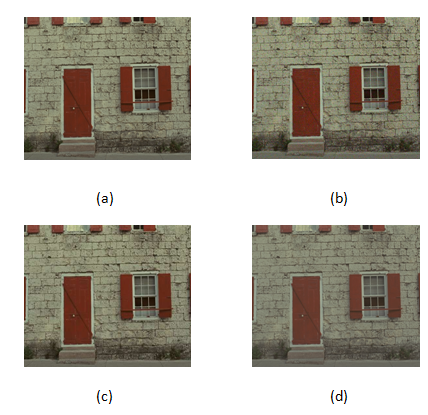} \caption{A reference image from TID 2008 database with some distortion type; (a) Reference image, (b) Additive noise in
color components, (c) JPEG compression, (d) Contrast change.}

\vspace{1.5ex}

\end{figure}

\begin{table*}[!t]
\vspace{1.5ex}
 \centering %
\begin{tabular}{|c|c|c|c|c|c|c|}
\hline
{\scriptsize Distortion type }  & {\scriptsize $RRED_{16}^{M16}$~\cite{key:9}  }  & {\scriptsize MSSIM~\cite{key:2} }  & {\scriptsize VIF~\cite{key:1} } & {\scriptsize PSNR }  & {\scriptsize WNISM~\cite{key:3} }  & {\scriptsize Our method}\tabularnewline
\hline
\hline
{\scriptsize Additive Gaussian noise}  & {\scriptsize 0.802}  & {\scriptsize 0.748}  & {\scriptsize 0.880}  & {\scriptsize 0.908}  & {\scriptsize 0.577}  & {\scriptsize 0.829}\tabularnewline
\hline
{\scriptsize Additive noise in}  & {\scriptsize 0.823}  & {\scriptsize 0.778}  & {\scriptsize 0.876}  & {\scriptsize 0.897}  & {\scriptsize 0.598}  & {\scriptsize 0.836}\tabularnewline
{\scriptsize color components is more intensive}  &  &  &  &  &  &  \tabularnewline
{\scriptsize than additive noise in the luminance component}  &  &  &  &  &  &  \tabularnewline
\hline
{\scriptsize Spatially correlated noise}  & {\scriptsize 0.839}  & {\scriptsize 0.760}  & {\scriptsize 0.862}  & {\scriptsize 0.910}  & {\scriptsize 0.567} & {\scriptsize 0.873}\tabularnewline
\hline
{\scriptsize Masked noise}  & {\scriptsize 0.827}  & {\scriptsize 0.816}  & {\scriptsize 0.819}  & {\scriptsize 0.850}  & {\scriptsize 0.609}  & {\scriptsize 0.842}\tabularnewline
\hline
{\scriptsize High frequency noise}  & {\scriptsize 0.893}  & {\scriptsize 0.822}  & {\scriptsize 0.903}  & {\scriptsize 0.902}  & {\scriptsize 0.811}  & {\scriptsize 0.871}\tabularnewline
\hline
{\scriptsize Impulse noise}  & {\scriptsize 0.712}  & {\scriptsize 0.625}  & {\scriptsize 0.815}  & {\scriptsize 0.823}  & {\scriptsize 0.583}  & {\scriptsize 0.829}\tabularnewline
\hline
{\scriptsize Quantization noise}  & {\scriptsize 0.829}  & {\scriptsize 0.757}  & {\scriptsize 0.777}  & {\scriptsize 0.852}  & {\scriptsize 0.615}  & {\scriptsize 0.824}\tabularnewline
\hline
{\scriptsize Gaussian blur}  & {\scriptsize 0.955}  & {\scriptsize 0.943}  & {\scriptsize 0.932}  & {\scriptsize 0.865}  & {\scriptsize 0.860}  & {\scriptsize 0.923}\tabularnewline
\hline
{\scriptsize Image denoising}  & {\scriptsize 0.946}  & {\scriptsize 0.915}  & {\scriptsize 0.914}  & {\scriptsize 0.934}  & {\scriptsize 0.858}  & {\scriptsize 0.903}\tabularnewline
\hline
{\scriptsize JPEG compression}  & {\scriptsize 0.929}  & {\scriptsize 0.931}  & {\scriptsize 0.898}  & {\scriptsize 0.868}  & {\scriptsize 0.829}  & {\scriptsize 0.830}\tabularnewline
\hline
{\scriptsize JPEG2000 compression}  & {\scriptsize 0.954}  & {\scriptsize 0.955}  & {\scriptsize 0.954}  & {\scriptsize 0.802}  & {\scriptsize 0.929}  & {\scriptsize 0.904}\tabularnewline
\hline
{\scriptsize JPEG transmission errors}  & {\scriptsize 0.863}  & {\scriptsize 0.867}  & {\scriptsize 0.836}  & {\scriptsize 0.745}  & {\scriptsize 0.859}  & {\scriptsize 0.852}\tabularnewline
\hline
{\scriptsize JPEG2000 transmission errors}  & {\scriptsize 0.715}  & {\scriptsize 0.836}  & {\scriptsize 0.838}  & {\scriptsize 0.827}  & {\scriptsize 0.686}  & {\scriptsize 0.712}\tabularnewline
\hline
{\scriptsize Non eccentricity pattern noise}  & {\scriptsize 0.703}  & {\scriptsize 0.721}  & {\scriptsize 0.756}  & {\scriptsize 0.534}  & {\scriptsize 0.427}  & {\scriptsize 0.544}\tabularnewline
\hline
{\scriptsize Local block-wise distortions of different intensity}  & {\scriptsize 0.798}  & {\scriptsize 0.751}  & {\scriptsize 0.816}  & {\scriptsize 0.609}  & {\scriptsize 0.573} & {\scriptsize 0.807}\tabularnewline
\hline
{\scriptsize Mean shift (intensity shift)}  & {\scriptsize 0.513}  & {\scriptsize 0.726}  & {\scriptsize 0.500}  & {\scriptsize 0.663}  & {\scriptsize 0.291}  & {\scriptsize 0.576}\tabularnewline
\hline
{\scriptsize Contrast change}  & {\scriptsize 0.513}  & {\scriptsize 0.632}  & {\scriptsize 0.808}  & {\scriptsize 0.521}  & {\scriptsize 0.563}  & {\scriptsize 0.528}\tabularnewline
\hline
{\scriptsize No.of scalars} & {\scriptsize L/36} & {\scriptsize L} & {\scriptsize L} & {\scriptsize L} & {\scriptsize 18} & {\scriptsize 54}\tabularnewline
\hline
\end{tabular}\caption{Comparative table of our method with the others by the Pearson's linear correlation coefficient (PLCC).\label{ls:tab}}

\vspace{1.5ex}

\end{table*}

\section{Experimental results}

In this section we evaluate the performances of the proposed method using TID 2008 database~\cite{key:14}.
This benchmark is intended for evaluation of full-reference image visual quality assessment metrics.
It allows to estimate how a given metric correlates with mean human
perception. For example, in accordance with TID 2008, Spearman correlation
between the metric PSNR (Peak Signal to Noise Ratio)
and mean human perception (MOS, Mean Opinion Score) is 0.525.
TID 2008 contains 25 reference images and 1700 distorted images (25
reference images with 17 types of distortions and 4 levels of distortions). These
distortions type are : Additive Gaussian noise, Additive noise in
color components, Spatially correlated noise, Masked noise, High frequency
noise, Impulse noise, Quantization noise, Gaussian blur, Image denoising, JPEG compression, JPEG2000 compression, JPEG transmission errors, JPEG2000 transmission errors, Non eccentricity pattern noise, Local
block-wise distortions of different intensity, Mean shift (intensity shift), Contrast change. Figure 3 shows one reference image from the TID 2008 benchmark with three distortion types.\\
The quality of each image in TID 2008 has been graded by the Mean Opinion Score(MOS) derived from psychophysical experiments.
We use here the Spearman rank correlation coefficient to estimate the correlation between the MOS and its prediction according to the proposed metric, it is given is :

\begin{equation}
SRCC=1-\frac{6\sum_{i=1}^{^{N}}d_{i}^{2}}{N(N^{2}-1)},
\end{equation}

Where $d_{i}$ is the difference between the $i^{th}$ image rank in
the subjective and objective evaluations. The Spearman rank correlation coefficient (SRCC) is a non-parametric
correlation metric, independent of any monotonic nonlinear mapping
between the subjective and the objective score.\\
We also use the Pearson's linear correlation coefficient (PLCC) defined by :

\begin{equation}
PLCC=\frac {\sum_{i=1}^{n}(s_i-\bar{s})(x_i-\bar{x})}{\sqrt {\sum_{i=1}^{n}(s_i-\bar{s})} \sqrt{\sum_{i=1}^{n}(x_i-\bar{x})}},
\end{equation}
 Where $s_i$ is the subjective score of the $i^{th}$ image, $x_i$ is the objective score of the $i^{th}$ image
 and ($\bar{s}$, $\bar{x}$) denote respectively the average of ($s$, $x$).

The predicted MOS are computed from the values generated by the objective
measure. We use a non-linear function proposed by the Video
Quality Expert Group (VQEG) Phase I FR-TV ~\cite{key:12} with five parameters.
The expression of this logistic function is given by:
\begin{equation}
\mathrm{Quality}(x)=\beta_{1}\mathrm{logistic}\left(\beta_{2},x-\beta_{3}\right)+\beta_{4}x+\beta_{5}.
\end{equation}

\begin{equation}
\mathrm{logistic}\left(\tau,x\right)=\frac{1}{2}-\frac{1}{1+\exp\left(\tau x\right)}.
\end{equation}

Table I and II shows respectively the spearman
rank correlation and the pearson's linear correlation coefficient for each types of distortion in the TID 2008 database for our method, three FR and two RR methods for comparative purposes. \\
Additionally, the number of  features extracted from the original image by each method is reported in the row denoted by \textquotedblleft{}number of scalars\textquotedblright{} in both tables in order to have a fair comparison (L denotes the image size).
 We remark that the information quantity used for our method is smaller than the others, except for the WNISM
 method.  As reported, only 54 features (six subbands and for each subband we have nine parameters) are sufficient to reach a good level of performances.
Table I shows that our method offers a good trade off between the existing methods for a great variety of distortion types.
For example, in the case of the gaussian blur, the quality score is higher than WNSIM and VIF and smaller than the others.
Our approach is also shown a highly efficient for the distortion type that attack the color components, such as Additive noise in color components.
 If we compare our approach with the WNISM, the results show the superiority of our method in the great majority of cases (12 times out of 15).
 WNSIM performs better only in the case of JPEG compression, JPEG transmission errors and contrast change artefacts.

Table II, report the PLCC values. As we can see, the correlation coefficients values of our method are close to the one obtained with the full-reference methods. And for the reduced-reference methods is clearly the superiority of our method against WNISM, also for the RRED method, the proposed metric outperforms for eight distortions.
However, the proposed measure fails for the Non eccentricity pattern noise distortion.

For each type of distortion under evaluation the PLCC and SRCC are computed from the scatter plots, where each point in the plots represent one test image.
The vertical axe is the MOS with the psychophysical experiments and the horizontal axe represent the values of the corresponding distortion measure.
If the prediction is perfect, then the point should lie on the line.

\begin{figure}[htb]
\vspace{1.5ex}
 \centering \includegraphics[scale=0.6]{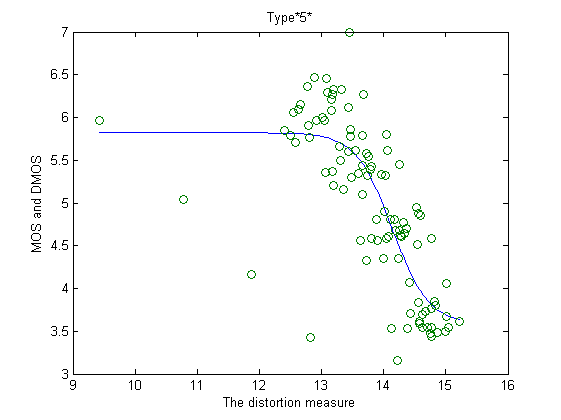} \caption{The scatter plot of the distortion Masked noise in TID 2008
database.}

\vspace{1.5ex}

\end{figure}

Figure 4 shows the scatter plots for the distortion type named \textquotedblleft Masked noise\textquotedblright with our method.
We can observe that a great number of images the values is closed to the line and that some points are placed on the line.

\begin{figure}[htb]
\vspace{1.5ex}
 \centering \includegraphics[scale=0.6]{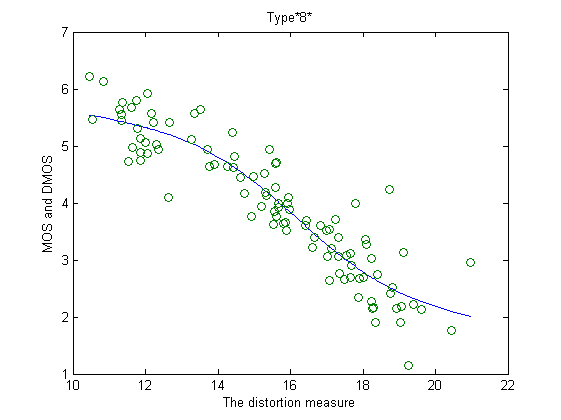} \caption{The scatter plot of the distortion Quantization noise in TID 2008
database.}

\vspace{1.5ex}

\end{figure}

Figure 5 illustrates the same type of results for the distortion \textquotedblleft Quantization noise\textquotedblright in TID 2008.
We can observe that for this type of distortion the points are more closed to the line, which puts in evidence the adequacy of the proposed metric in this situation.

\section{Conclusion}
In this paper, we proposed a RR measure based on Multivariate Generalized Gaussian Distribution.
The MGGs Distribution is intended to handle the dependencies between the color components.
The Kullback-Leibler divergence between two MGGD is used to compute the visual quality degradation.
Results of an extensive comparative evaluation show that the method is very effective for a broad range of distortion types.
Furthermore  just a limited  quantity information from the reference image is required in order to compute the quality assessment measure.
With only 54 parameters from the reference image the algorithm achieves a performance which is nearly as good as the best performing full reference quality assessment methods.
Future work will concern other color spaces and an extension to video quality assessment.





%

\end{document}